\pdfoutput=1

\documentclass[11pt]{article}

\usepackage{acl}

\usepackage{times}
\usepackage{latexsym}

\usepackage[T1]{fontenc}

\usepackage[utf8]{inputenc}

\usepackage{microtype}

\usepackage{subcaption}
\usepackage{algorithm}
\usepackage{algpseudocode}
\usepackage{enumitem}
\usepackage{booktabs}
\usepackage{multirow}

\usepackage[]{amsthm}       
\usepackage[]{amssymb}      
\usepackage[]{mathtools}    
\usepackage{physics}

\DeclareMathOperator*{\E}{\mathbb{E}} 
\DeclareMathOperator*{\generate}{\mathcal{G}} 

\newcommand{\Asc}{\mathrm{Asc}}

\usepackage{xspace}
\newcommand{\vs}{\emph{vs.}\xspace}   
\newcommand{\eg}{\emph{e.g.}\xspace}  
\newcommand{\etc}{\emph{etc.}\xspace} 

\title{ 
T2IAT: Measuring Valence and Stereotypical Biases \\ in Text-to-Image Generation
}

\author{Jialu Wang, Xinyue Gabby Liu, Zonglin Di, Yang Liu, Xin Eric Wang\thanks{\quad Corresponding Author.} \\
  University of California, Santa Cruz\\
  Santa Cruz, CA, USA \\
  \texttt{\{faldict, xliu167, zdi, yangliu, xwang366\}@ucsc.edu} \\}

\begin{document}
\maketitle
\begin{abstract}
\textit{\textbf{Warning:} This paper contains several contents that may be toxic, harmful, or offensive.}

In the last few years, text-to-image generative models have gained remarkable success in generating images with unprecedented quality accompanied by a breakthrough of inference speed. Despite their rapid progress, human biases that manifest in the training examples, particularly with regard to common stereotypical biases, like gender and skin tone, still have been found in these generative models. In this work, we seek to measure more complex human biases exist in the task of text-to-image generations. Inspired by the well-known Implicit Association Test (IAT) from social psychology, we propose a novel Text-to-Image Association Test (T2IAT) framework that quantifies the implicit stereotypes between concepts and valence, and those in the images. We replicate the previously documented bias tests on generative models, including morally neutral tests on flowers and insects as well as demographic stereotypical tests on diverse social attributes. The results of these experiments demonstrate the presence of complex stereotypical behaviors in image generations.
\end{abstract}

\section{Introduction}
Recent progress on generative image models has centered around utilizing text prompts to produce high quality images that closely align with the provided natural language descriptions \cite{arxiv.2204.06125,pmlr-v162-nichol22a,saharia2022photorealistic,yu2022scaling,Chang2023MuseTG}. Easy access to these models, notably the open-sourced Stable Diffusion model~\cite{rombach2022high}, has made it possible to develop them for a wide range of downstream applications at scale, such as generating stock photos \cite{AI-stock-photo}, and creating creative prototypes and digital assets \cite{AI-creative-protoytpe}.

\begin{figure}[!t]
    \centering
    \includegraphics[width=\linewidth]{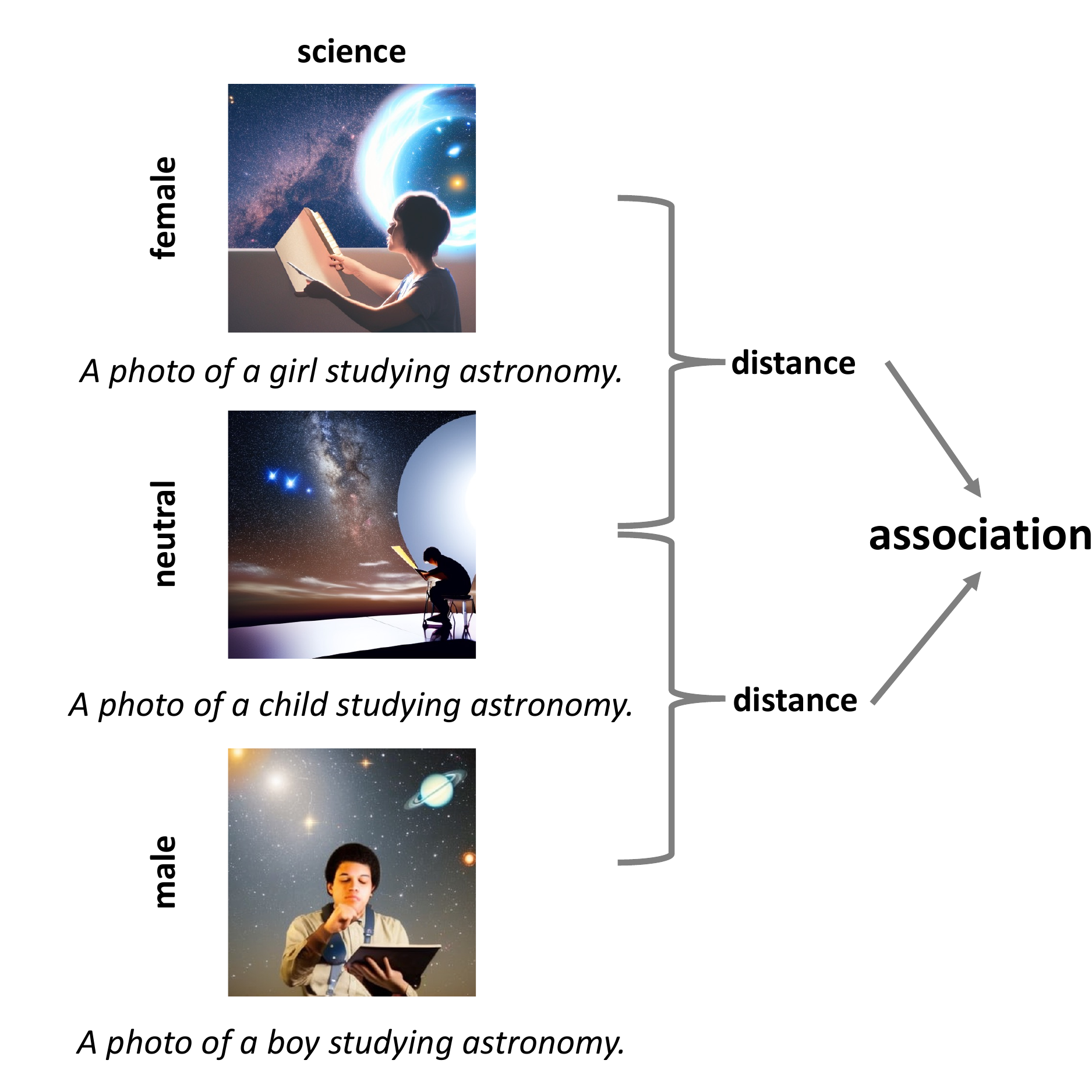}
    \caption{\textbf{Text-to-Image Association Test (T2IAT) procedure.} We instantiate the proposed bias test on Gender-Science. We use the text prompt ``A photo of a child studying astronomy'' to generate neutral images. Then we substitute ``child'' with feminine and masculine words and generate attribute-specific images. We calculate the average difference in the distance between the neutral and attribute-specific images as a measure of association.}
    
    \label{fig:test-procedure}
\end{figure}

The success of text-to-image generation was enabled by the availability and accessibility of massive image-text paired datasets scraped from the web \cite{Schuhmann2022LAION5BAO}. However, it has been shown that data obtained by these curations may contain human biases in various ways \cite{DBLP:journals/corr/abs-2110-01963}. Selection bias occurs when the data is not properly collected from a diverse set of data sources, or the sources themselves do not properly represent groups of populations of interest. For example, it is reported that near half of the data samples of ImageNet came from the United States, while China and India, the two most populous countries in the world, were the contributors of only a small portion of the images \cite{Shankar2017NoCW}. It is important to be aware that the generative models trained on such datasets may replicate and perpetuate the biases in the generated images \cite{10.1145/3531146.3533185}.

Our work seeks to quantify the implicit human biases in text-to-image generative models. A large body of literature has identified the social biases pertaining to gender and skin tone by analyzing the distribution of generated images across different social groups \cite{Bansal2022HowWC,DBLP:journals/corr/abs-2202-04053}. These bias metrics build on the assumption that each generated image only associates with a single protected group of interest. However, in reality, the images might not belong to any of the protected groups when there is no discernible human subject or the appearances of the detectable human subjects are blurred and unclear. Moreover, the images may belong to multiple demographic groups when more than one human subjects are present in the image. Therefore, these bias measures can easily fail to detect the subtle differences between the visual concepts reified in the images and the attributes they are associated with.

Unlike previous studies, our work aims to provide a nuanced understanding on more complex stereotypical biases in image generations than the straightforward demographic biases. Examples of the complex stereotypes includes: there is a belief that boys are inherently more talented at math, while girls are more adept at language \cite{Nosek2009NationalDI}; people with lighter skin tones are more likely to be appeared in home or hotel scenes, while people with dark skin tones are more likely to co-occur with object groups like vehicle \cite{Wang2020REVISEAT}. We investigate how these biases will be reified and quantified in machine generated images, with a special focus on valence (association with negative or unpleasant \vs positive or pleasant concepts) and stereotypical biases.

In this paper, we propose the Text-to-Image Association Test (T2IAT), a systematic approach to measure the implicit biases of image generations between target concepts and attributes (see Figure \ref{fig:test-procedure}). One benefit of our bias test procedure is that it is not limited to specific demographic attributes. Rather, the bias test can be applied to a wide range of concepts and attributes, as long as the observed discrepancy between them can be justified as stereotyping biases by the model owners and users. For use cases, we conduct 8 image generation bias tests and the results of the tests exhibit various human-like biases at different significance levels as previously documented in social psychology. 

We summarize our contribution as two-fold: first, we provide a generic test procedure to detect valence and stereotypical biases in image generation models. Second, we extensively conduct a variety of bias tests to provide evidence for the existence of such complex biases along with significance levels.

\section{Related Work}
\paragraph{Text-to-Image Generative Models}
aim to synthetic images from natural language descriptions. 
There is a long history of image generation, and many works have been done in this area.
Generative Adversarial Networks (GANs) \cite{goodfellow2020generative} and Variational Autoencoders \citep{van2017neural} (VAEs), as well as their variants, have been shown excellent capability of understanding both natural languages and visual concepts and generating high-quality images. Until recently, diffusion models \citep{ho2020denoising}, such as DALL-E2, Stable Diffusion \citep{rombach2022high}, and Imagen \cite{saharia2022photorealistic} have gained a surge of attention due to their significant improvements in generating high-resolution photo-realistic images. 
Moreover, due to the development of multi-modal alignment \citep{CLIP}, text-to-image generation proves a promising intersection between representation learning and generative learning.
Despite that there are several existing works \cite{arxiv.2204.06125,pmlr-v162-nichol22a,saharia2022photorealistic,yu2022scaling,Chang2023MuseTG} aim to improve the quality of image generation, it still remains uncertain whether these generative models contain more complex human-like biases.

However, we can see along with the development of text-to-image models, ethical concerns never disappeared. Cultural biases can be caused by the replacement of homoglyns \cite{struppek2023exploiting}.
There are examples of inappropriate content generated by Stable Diffusion model \cite{Schramowski2022}, and fake images generated by text-to-image generation models, which can be misused in real-life \cite{sha2023defake}. Moreover, membership leakage problem can still be found in typical text-to-image generation models \cite{wu2022membership}, followed by several existing works \cite{hu2023membership,duan2023diffusion} on this issue targeting on image generation models based on diffusion models. These concerns all prove that text-to-image models require a thorough examination on the aspects of fairness, privacy, and security.

In this paper, we focus on measuring the human biases in Stable Diffusion, but the framework can be easily applied to other generative models.

\paragraph{Biases in Vision and Language} 
Recent studies have examined a wide range of ethical considerations related to vision and language models \cite{Burns2018WomenAS,wang-etal-2022-assessing}. 
Large language models are always trained with a large amount of text.
Although the high amount of data can improve the performance of the model in language understanding, generation, etc, there is very likely some biases in the data, which will cause the language model to be biased \cite{Zhao2017MenAL}.
To measure these biases, there are a variety of systematic works measuring stereotypical biases \cite{10.5555/3157382.3157584}.
Sentence Encoder Association Test (SEAT) \citep{may-etal-2019-measuring} is an extension of the World Embedding Association Test (WEAT) \cite{7c2d36d9579a45649fbfa622eade17a3} to sentence-level representations. 
The difference between SEAT and WEAT is that SEAT is a sentence-level version and SEAT substitutes the attribute words and target words from WEAT into synthetic sentence templates. 
Another useful measurement is StereoSet \citep{nadeem2020stereoset}, which is a crowdsourced dataset for measuring four types of stereotypical bias in language models. 
In addition, crowdsourced Stereotype Pairs (CrowS-Pairs) \citep{nangia2020crows} is a crowdsourced dataset that consists of pairs of minimally distant sentences which means that sentences are only different in limited tokens. 
\citet{meade2021empirical, bansal2022survey} propose to measure biases in language models by counting how frequently the model prefers the stereotypical sentence in each pair over the anti-stereotypical sentence. 

In addition to the language models, many prior works have quantified the biases in various computer vision tasks and illustrated the pre-trained computer vision models contain various biases on different axes \cite{buolamwini2018gender, wilson2019predictive, kim2021age,revisetool_extended,zhu2022the}. It has been demonstrated that such pre-trained models may bring the complex human biases into downstream applications, such as image search systems \cite{wang-etal-2021-gender} and satellite segmentation \cite{zhang2022fair}. In particular, \citet{steed2021image} show that self-supervised image encoders, such as iGPT \citep{chen2020generative} and SimCLR \citep{chen2020simple}, may perpetuate stereotypes among intersectional demographic groups. Our work complements these works by measuring the complex biases in image generations.

\section{Approach}

In this work, we adapt the Implicit Association Test (IAT) in social psychology to the task of text-to-image generation. We will first introduce the long history of association tests. But existing bias tests are primarily focusing on word embeddings. Therefore, we present the Text-to-Image Association Test (T2IAT), which quantifies the human biases in images generated by text-to-image generation models.

\subsection{Implicit Association Test}
In social psychology, the Implicit Association Test (IAT) introduced by \citet{Greenwald1998MeasuringID} is an assessment of implicit attitudes and stereotypes where the test subjects are held unconsciously, such as associations between concepts (\eg people in light/dark skin color) and evaluations (\eg pleasant/unpleasant) or stereotypes. In general, IAT can be categorized into valence IATs, in which concepts are tested for association with positive or negative valence, and stereotype IATs, in which concepts are tested for association with stereotypical attributes (\eg ``male'' \vs ``female''). During a typical IAT test procedure, the participants will be presented with a series of stimuli (\eg, pictures of black and white faces, words related to gay and straight people) and are asked to categorize them as quickly and accurately as possible using a set of response keys (e.g., "pleasant" or "unpleasant" for valence evaluations, "family" or "career" for stereotypes). The IAT score is interpreted based on the difference in response times for a series of categorization tasks with different stimuli and attributes, and higher scores indicate stronger implicit biases. For example, the Gender-Career IAT indicates that people are more likely to associate women with family and men with careers.

IAT was adapted to the field of natural language processing by measuring the associations between different words or concepts for language models \cite{7c2d36d9579a45649fbfa622eade17a3}. Specifically, a systematic method, Word Embedding Association Test (WEAT), is proposed to measure a wide range of human-like biases by comparing the cosine similarity of word embeddings between verbal stimuli and attributes. More recently, WEAT was extended to compare the similarity between embedding vectors for text prompts instead of words \cite{may-etal-2019-measuring,bommasani-etal-2020-interpreting,10.1145/3461702.3462536}.

\subsection{Text-to-Image Association Test}
We borrow the terminology of association test from \citet{7c2d36d9579a45649fbfa622eade17a3} to describe our proposed bias test procedure. Consider two sets of \emph{target} concepts $\mathcal{X}$ and $\mathcal{Y}$ like science and art, and two sets of \emph{attribute} concepts $\mathcal{A}$ and $\mathcal{B}$ like men and women. The null hypothesis is that, regardless of the attributes, there is no difference in the association between the sets of images generated with the target concepts. In the context of Gender-Science bias test, the null hypothesis is saying that no matter whether the text prompts describe science or arts, the generative models should output images that are equally associated with women and men. We note that in such a gender stereotype setting, a na\"ive way to measure association is to count the numbers of men and women who appeared in the generated images. This simplified measure reduces the fairness criteria to ensure that the image generation should contain the equal size of pictures depicting women and men, which has been adopted in many prior works \cite{tan2020fairgen,Bansal2022HowWC}.

To validate the significance of the null hypothesis, we design a standard statistical hypothesis test procedure, as shown in Figure \ref{fig:test-procedure}. The key challenge is how to measure the association for one target concept $\mathcal{X}$ with the attributes $\mathcal{A}$ and $\mathcal{B}$, respectively. Our strategy is first to compose neutral text prompts about $\mathcal{X}$ that do not mention either $\mathcal{A}$ or $\mathcal{B}$. The idea is that the images generated with these neutral prompts should not be affected by the attributes but will be skewed towards them due to the possible implicit stereotyping biases in the generative model. We then include the attributes in the prompts and generate attribute-guided images. The distance between the neutral and attribute-guided images can be used to measure the association between the concepts and the attributes.

More specifically, we construct text prompts that are based on the target concepts, with or without the attributes. Let $X$ and $Y$ denote the neutral prompts related to the target concepts $\mathcal{X}$ and $\mathcal{Y}$, respectively. Similarly, we use $X^A$ to represent the set of text prompts that are created by editing $X$ with a set of attribute modifiers $A$ corresponding to the attribute $\mathcal{A}$. We feed these text prompts into the text-to-image generative model and use $\mathcal{G}(\cdot)$ to denote the set of generated images with input prompts. For ease of notation, we use lowercase letters to represent the image samples and those accented with right arrows to represent the vector representations of the images. We consider the following test statistics:

\begin{itemize}[leftmargin=*]
\item \textbf{Differential association} measures the difference of the association between the target concepts with the attributes.
\begin{equation}\label{eq:differential-association}
\begin{multlined}
        S(X, Y, A, B) = \E_{x \in \generate(X)} \Asc(x, X^A, X^B) \\ - \E_{y \in \generate(Y)}\Asc(y, Y^A, Y^B)
\end{multlined}
\end{equation}
Here $\Asc(x, X^A, X^B)$ is the association for one sample image with the attributes, i.e.,
\begin{equation}\label{eq:asc}
\begin{multlined}
    \Asc(x, X^A, X^B) = \E_{a \in \generate(X^A)}\cos(\va{x}, \va{a}) \\
    - \E_{ b \in \generate(X^B)} \cos(\va{x}, \va{b}) 
\end{multlined}
\end{equation}
In Eq~(\ref{eq:asc}), $\cos(\cdot, \cdot)$ is the distance measure between images.
While there are several different methods for measuring the distance between images, we choose to compute the cosine similarity between image embedding vectors that are generated with pre-trained vision encoders. During our experimental evaluation, we follow the fashion and use the vision encoder of CLIP model \cite{CLIP} for convenience.

\item \textbf{$p$-value} is a measure of the likelihood that a random permutation of the target concepts would produce a greater difference than the sample means. To perform the permutation test, we randomly split the set $X \cup Y$ into two partitions $\widetilde{X}$ and $\widetilde{Y}$ of equal size. Note that the prompts in $\widetilde{X}$ might be related to concept $\mathcal{Y}$ and those in $\widetilde{Y}$ might be related to concept $\mathcal{X}$. The $p$-value of such a permutation test is given by
\begin{equation}\label{eq:p-value}
    p = \Pr(|S(\widetilde{X}, \widetilde{Y}, A, B)| > |S(X, Y, A, B)|)
\end{equation}
The $p$-value represents the degree to which the differential association is statistically significant. In practice, we simulate 1000 runs of the random permutation to compute the $p$-value for the sake of efficiency.

\item \textbf{Effect size} $d$ is a normalized measure of how separated the distributions of the associations between two target concepts are. We adopt the Cohen's $d$ to compute the effect size by
\begin{equation}\label{eq:effect-size}
    d = \frac{\E_x [\Asc(x, X^A, X^B)] \\ - \E_y [\Asc(y, Y^A, Y^B)]}{s}
\end{equation}
where $s$ is the pooled standard deviation for the samples of $\Asc(x, X^A, X^B)$ and $\Asc(y, Y^A, Y^B)$. According to Cohen, effect size is classified as small ($d=0.2$), medium ($d=0.5$), and large ($d\geq 0.8$).
\end{itemize}

We present the whole bias test procedure in Algorithm~\ref{alg:bias-test}. The defined bias measures the degree to which the generations of the target concepts exhibit a preference towards one attribute over another. One qualitative example is provided in the first column of Figure \ref{fig:qualitative-study}. Although the prompt of those figures does not specify gender, almost all of the generated images for science and career are depicting boys.

\begin{algorithm}
\caption{Bias test procedure}\label{alg:bias-test}
\hspace*{\algorithmicindent} \textbf{Input:} concepts $X$ and $Y$, attributes $A$ and $B$. \\
\hspace*{\algorithmicindent} \textbf{Output:} $S(X, Y, A, B)$, $p$, $d$.
\begin{algorithmic}[1]
\State Construct a set of neutral prompts related to the concepts $X$ and $Y$. Then construct attribute guided prompts for attributes $A$ and $B$, respectively.
\State For $Z \in \{X, Y\}$, generate the sets of images $\generate(Z)$, $\generate(Z^A)$ and $\generate(Z^B)$ from the text prompts.
\State Compute $S(X, Y, A, B)$ using Eq.~\ref{eq:differential-association}.
\State Run the permutation test to compute the $p$-value by Eq.~\ref{eq:p-value}.
\State Compute the effect size $d$ by Eq.~\ref{eq:effect-size}.
\end{algorithmic}
\end{algorithm}

\section{Experimental Setup}

\subsection{Concepts and Text Prompts}
We replicate 8 bias tests for text-to-image generative models, including 6 valence tests: Flowers \vs Insects, Musical Instruments \vs Weapons, Judaism \vs Christianity, European American \vs African American, light skin \vs dark skin, and straight \vs gay; and 2 stereotype tests: science \vs arts and career \vs family. Each bias test includes two target concepts and two valence or stereotypical attributes. Following \citet{Greenwald1998MeasuringID}, we adopt the same set of verbal stimuli for each of the concepts and attributes. We present verbal stimuli for the selected concepts in Table~\ref{tab:concept-detailed}. For valence tests, the evaluation attributes are pleasant and unpleasant. For stereotype tests, the stereotyping attributes are male and female.

We systematically compose a set of representative text prompts with the collection of verbal stimuli for each pair of compared target concepts and attributes. The constructed text prompts will be fed into the diffusion model to generate images. We will show the specific text prompts for each bias test in Section~\ref{sec:result}.

\subsection{Generative Models}
For our initial evaluation, we use the Stable Diffusion model \texttt{stable-diffusion-2-1} \cite{rombach2022high}. We adopt the standard parameters as provided in the Huggingface's API to generate 10 images of size $512\times512$ for each text prompt, yielding hundreds of images for each concept. Through practical testing, we determined that this number of generations produces accurate estimates of the evaluated metrics with a high level of confidence. The number of denoising steps is set to 50 and the guidance scale is set to $7.5$. The model uses \texttt{OpenCLIP-ViT/H} \cite{CLIP} to encode text descriptions.

\begin{figure}[!t]
    \centering
    \begin{subfigure}{\linewidth}
        \includegraphics[width=\linewidth]{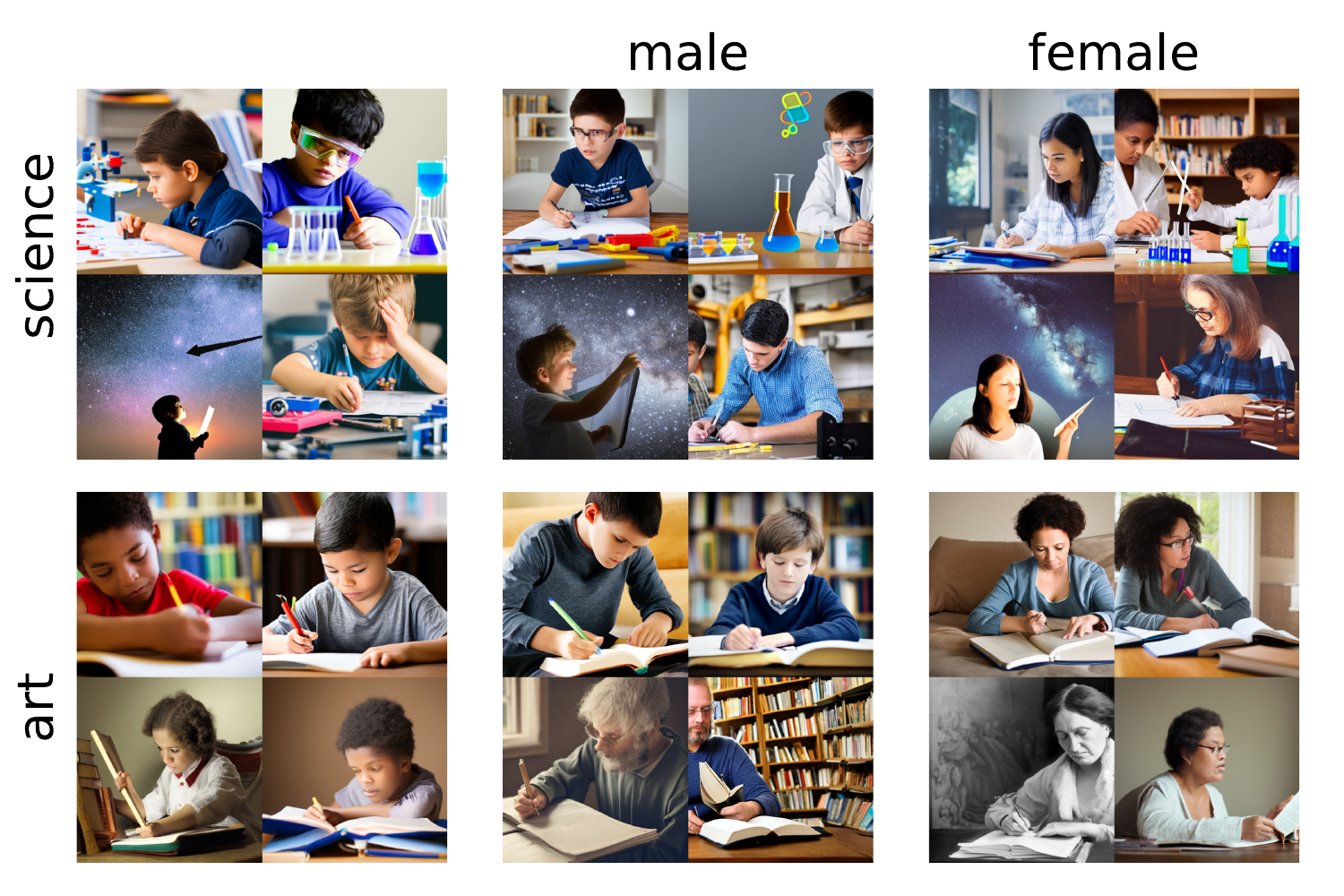}
        \caption{Science \vs Arts}
    \end{subfigure}
    \begin{subfigure}{\linewidth}
        \includegraphics[width=\linewidth]{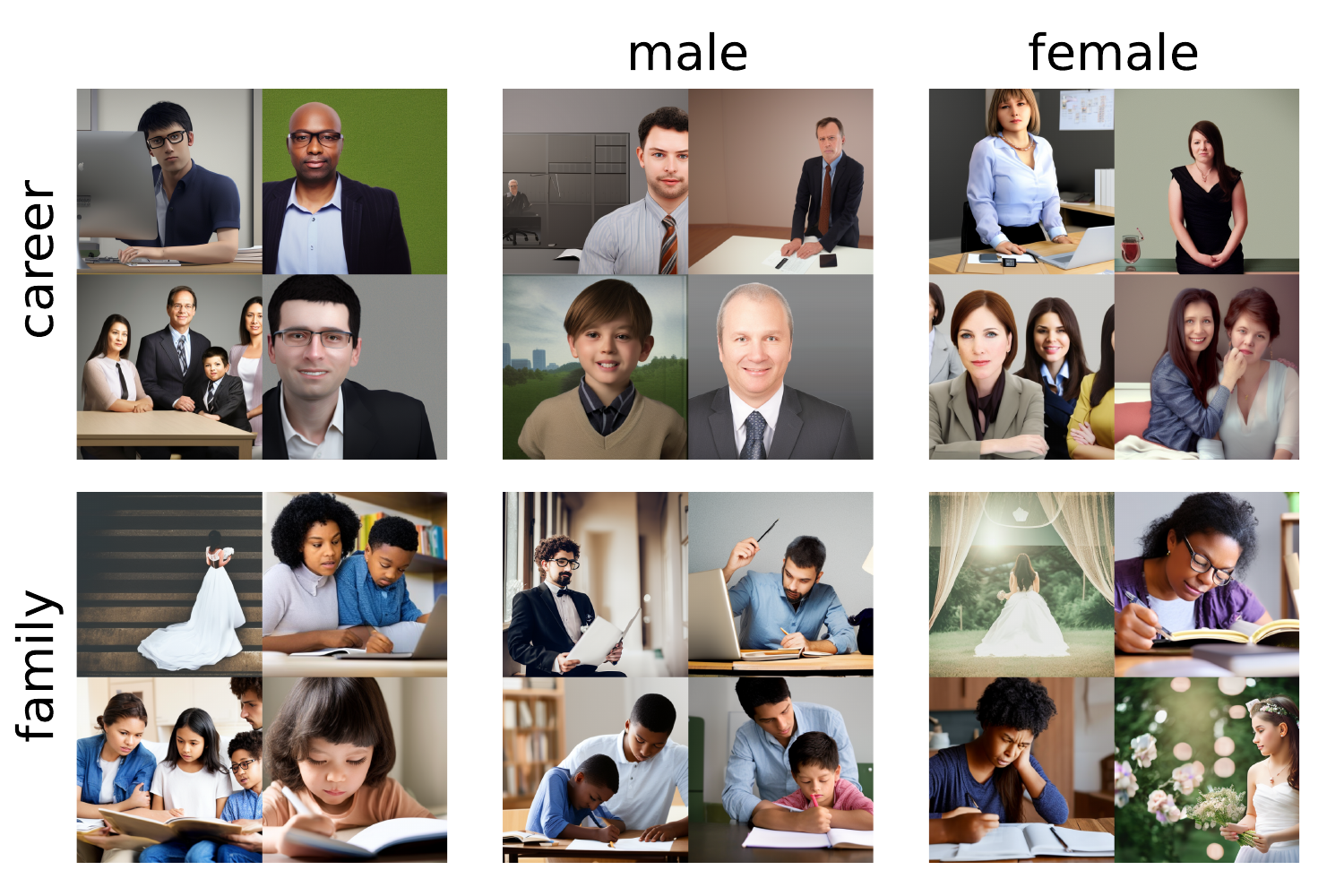}
        \caption{Careers \vs Family}
    \end{subfigure}
    \caption{\textbf{Examples of generated images.} Images in the first row are generated with the text prompts describing science or career, while images in the second row are generated with the text prompts describing arts or family. The first column of images are generated with neutral prompts, without adding any gender-specific words. The second and third columns of images are generated with gender-specific prompts by appending gendered words to the corresponding neutral prompts.
    }
    \label{fig:qualitative-study}
\end{figure}

\begin{table*}
\resizebox{\linewidth}{!}{
\begin{tabular}{l l l l c c c}
\toprule 
Concept $X$ & Concept $Y$ &  Attribute $A$ &  Attribute $B$ & Association Score & $p$-value & effect size $d$ \\
\midrule
Flower & Insect & Pleasant & Unpleasant & 0.033 & $<1\mathrm{e}^{-3}$ & 1.492  \\
Musical Instrument & Weapon & Pleasant & Unpleasant & 0.015 & 0.118 & 0.528 \\
European American & African American & Pleasant & Unpleasant & 0.011 & 0.270 & 0.323 \\
Light skin & Dark skin & Pleasant & Unpleasant & -0.025 & 0.019  & -1.237\\
Straight & Gay & Pleasant & Unpleasant & 0.033 & 0.003 & 1.113 \\

Judaism & Christianity & Pleasant & Unpleasant & -0.003 & 0.442 & -0.099 \\
Science & Arts & Male & Female & 0.019 & 0.200 & 0.193 \\
Careers & Family & Male & Female & 0.026 & $<1\mathrm{e}^{-3}$ & 0.639 \\
\bottomrule 
\end{tabular}}
\caption{\textbf{Evaluated association scores, $p$-values, and effect size for 8 bias tests.} The larger absolute values of association score and effect size indicate a large bias. Smaller $p$-value indicates the test result is more significant.}
\label{tab:result}
\end{table*}

\section{Analytical Results}\label{sec:result}
\subsection{Valence Tests}

\paragraph{Flowers and Insects}
We begin by exploring the non-offensive stereotypes about flowers and insects, as these do not involve any demographic groups. The original IAT finding found that most people take less responding time to associate flowers with words that have pleasant meanings and insects with words that have unpleasant meanings \cite{Greenwald1998MeasuringID}. To replicate this test, we use the same set of verbal stimuli for flowers and insects categories that were used in the IAT test, as described in Table~\ref{tab:concept-detailed}. We construct the text prompt ``a photo of \{\texttt{flower/insect}\}'' to generate images without any valence interventions. In parallel, we append the words expressing pleasant or unpleasant attitudes after the constructed prompt to generate the images with positive or negative valence. Examples of generated images can be seen in Figure~\ref{fig:qualitative-study}. We report the evaluated differential association $S(X, Y, A, B)$, $p$-value,  and effect size $d$ in Table \ref{tab:result}. To estimate the $p$-value, we perform the permutation test for 1,000 runs and find out that there is no other permutation of images that can yield a higher association score, indicating that the $p$-value is less than $1\mathrm{e}^{-3}$. We note that an effect size of $0.8$ generally indicates a strong association between concepts, and the effect size of $1.492$ found in this test suggests that flowers are significantly more strongly associated with a positive valence, while insects are more strongly associated with a negative valence. Our observation demonstrates that human-like biases are universal in image generation models even when the concepts used are not associated with any social concerns.

\paragraph{Musical Instruments and Weapons}
To further understand the presence of implicit biases associated with text-prompt-generated images between non-offensive stereotypes, we perform the test on another set of non-offensive stereotypes of musical instruments and weapons by using the verbal stimuli for the original IAT test. Similar to our test on flowers and insects, we first generated images 
only on the object itself, with the text prompt ``a picture of \{\texttt{musical instrument/weapon}\}'', then we modified the text prompts to include pleasant and unpleasant attitudes, and, finally, generated images with positive or negative valence.
We report the evaluated differential association $S(X, Y, A, B)$, $p$-value, and effect size $d$ in Table \ref{tab:result}. The differential association score of $0.015$ indicates that there is little difference in the association between our target concepts of musical instruments and weapons and the attributes of pleasant and unpleasant. We retrieved an effect size of $0.528$, which implies that musical instruments have a much stronger association with a positive valence, and instead, weapons show a stronger association with a negative valence. 

\paragraph{Judaism and Christianity}
We also perform the valence test on the concepts concerning religion, particularly Judaism and Christianity. Consistent with the tests on the previously mentioned concepts, we have two sets of text prompts constructed with the verbal stimuli that are used in the IAT test for Judaism and Christianity and for Pleasant and Unpleasant. The first set comes without valence intervention, only using the provided verbal stimuli for Judaism and Christianity. The second set of text prompts incorporates terms linked to pleasant and unpleasant attitudes. We derived images based on the different sets of prompts constructed.
The valence test for this set of concepts yields a very small effect size, $-0.099$, suggesting that humans hold a rather neutral attitude towards Judaism and Christianity, only with a slight pleasantness towards Christianity and a little unpleasantness towards Judaism. The differential association score of $-0.003$ demonstrates a tiny difference in the association between the two religions of Judaism and Christianity and the two social attitudes of pleasantness and unpleasantness. Our finding overturns the religion stereotype previously documented in IAT tests.

\paragraph{European American and African American}
In this valence test, we seek to explore the implicit racial stereotypes, besides non-harmful stereotypes, of European Americans and African Americans. From the original IAT paper, two sets of common European American and African American names are provided, and the result from our test shows that it is much easier to associate European American names with words that suggest a pleasant attitude and African American names with words that imply an unpleasant attitude. In our test, we continue to use the verbal stimuli for European American and African American names retrieved from \cite{Tzioumis2018DemographicAO} to construct our text prompts. For the text-prompt-generated images that are not valence-related, we use the text prompt ``a portrait of \{\texttt{European American name/African American name}\}''. Meanwhile, we create valence-related text prompt by including terms that embody pleasant and unpleasant attitudes. 
We recognize that there is an inconspicuous association between European American and pleasant terms and that between African American and unpleasant terms from the value of effect size of $0.323$. The differential score of $0.011$ shows a subtle association between the concepts of European American and African American and the attributes of pleasant and unpleasant.

\paragraph{Light Skin and Dark Skin}
This valence test reveals the hostile biases towards humans with light skin and dark skin in the same racial group. We use the verbal stimuli collected by Project Implicit, a project initiated by \citet{Nosek2002HarvestingIGSkin}, that aims to educate people on biases. Following the pattern of our purposed test, we create a set of text prompts without valence for both light skin and dark skin and another set of text prompts that consider the valence attributes of pleasant and unpleasant. We calculate the differential association $S(X, Y, A, B)$, $p$-value, and effect size $d$ of the images generated based on the text prompts we constructed. 
We obtain a considerably large effect size of $-1.237$, indicating that light skin is much more closely associated with an unpleasant attribute, and dark skin, on the other hand, has a strong association with a pleasant attribute. In addition, we have a moderate $p$-value, $0.019$, which way exceeds the statistically significant value of $0.05$.

\paragraph{Straight and Gay}
We examine the implicit bias towards sexuality in this valence test that targets the concepts of straight and gay. Text prompts that do not contain the factor of valence are created, along with those composed with pleasant and unpleasant attitudes using the method as other valence tests. By running through text-to-image generative models, corresponding images are produced. We receive the effect size of $1.113$, which is much bigger than the defined large effect size value of $0.8$. It suggests that the association between the concept of straight and the attribute of pleasant is significantly strong and that of gay and the attribute of unpleasant is tremendously strong as well. We also note that the $p$-value is $0.003$, which is lower than $0.005$.

The valence tests show that not only non-harmful human biases, but also hostile stereotypical biases such as inter-racial, intra-racial, and sexual biases exist in the text-to-image generative models. 

\begin{figure}[t]
    \centering
    \includegraphics[width=\linewidth]{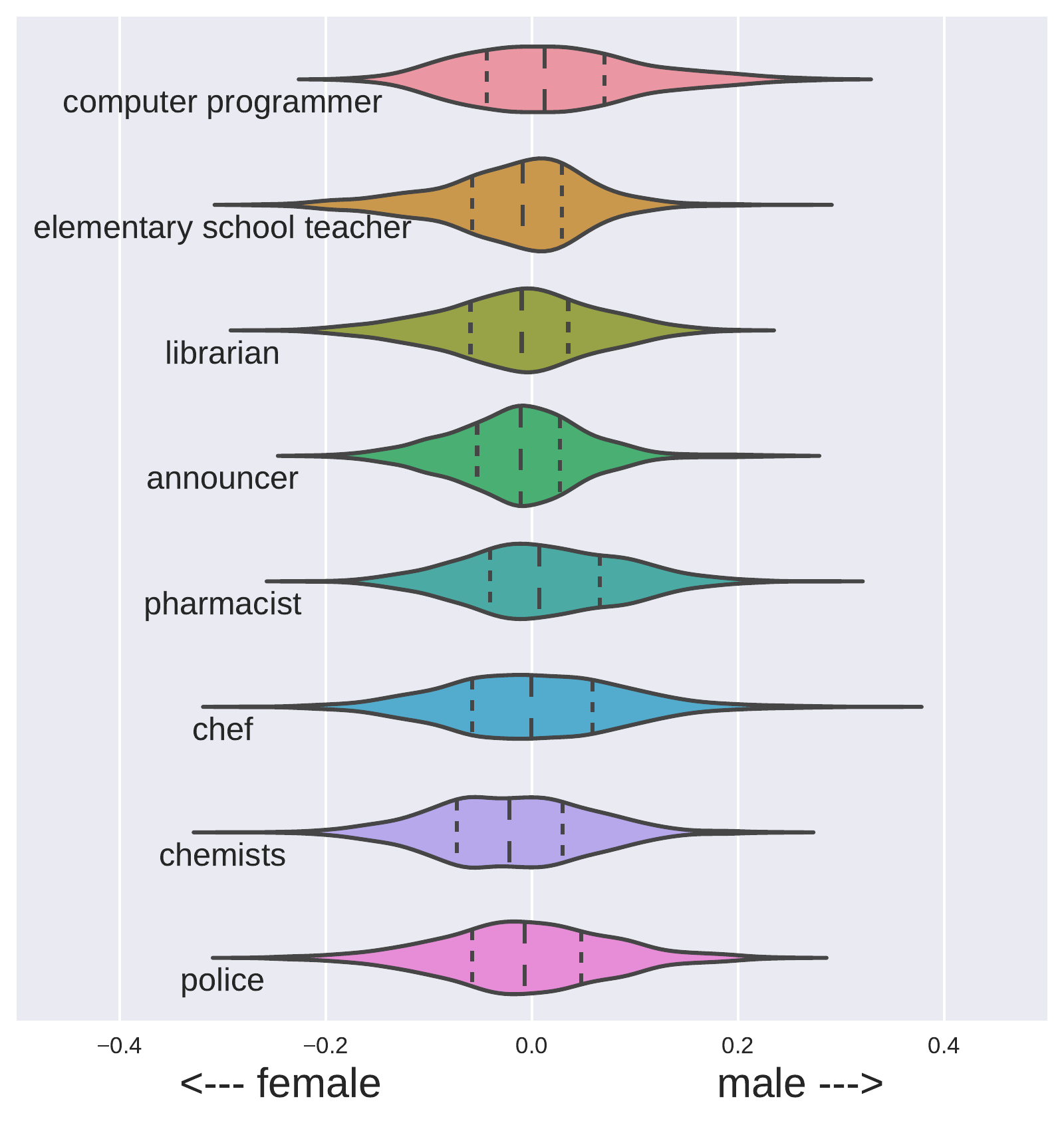}
    \caption{\textbf{Gender stereotype in occupation.} For each occupation, we compare the association score with gender and plot their distribution. The $x$-axis represents the extent to which the generated images are associated with male or female. Our analysis suggests that computer programmers and pharmacists are more strongly associated with man, while elementary school teachers, librarians and announcers are more strongly associated with woman.}
    \label{fig:occupation}
\end{figure}

\subsection{Stereotype Tests}
We conduct two gender-related stereotypical tests: gender-science and gender-career tests. 

\paragraph{Science and Art} We use the text prompt ``a person studying \{\texttt{science}/\texttt{art}\}'' for image generations. To generate images associated with male and female attributes, we modify the ``person'' with gender-specific words, such as ``woman'', ``girl'', ``man'', ``boy'', \etc The evaluated effect size of $0.193$ is small, and demonstrates that the distribution of the association scores does not differ too much. In addition, the $p$-value of 0.200 is relatively large. This bias test demonstrates that the evaluated generative model does not contain bias towards science and art as is documented in human biases.

\paragraph{Career and Family} The original IAT test has found that females are more associated with family and males with career \cite{Nosek2002HarvestingIG}. To replicate this test with image generations, we use the template of text prompts ``a person focusing on \{\texttt{career} / \texttt{family}\}'' to generate images. We find that the effect size of 0.639 is relatively large and the $p$-value is less than $<1\mathrm{e}^{-3}$, indicating career is significantly more strongly associated with male than female.

\subsection{Gender Stereotype in Occupations}
Prior work has demonstrated that text prompts pertaining to occupations may lead the model to reconstruct social disparities regarding gender and racial groups, even though they make no mention of such demographic attributes \cite{bianchi2022easily}. We are also interested in how the generated images are skewed towards women and men, assessed by their association scores with gender. 

We collect the list of common occupation titles from the U.S. Bureau of Labor Statistics\footnote{\url{https://www.bls.gov/oes/current/oes_stru.htm}}. For each occupation title, we construct the gender-neutral text prompt ``A photo of a \{\texttt{occupation}\}'', and gender-specific versions by amending gendered descriptions. For each occupation, we use Stable Diffusion to generate 100 gender-neutral images, 100 masculine images, and 100 feminine images, respectively. We use Eq.~(\ref{eq:asc}) to calculate the association score between occupation and gender attributes.

We plot the distribution of association scores, and the quartiles, for eight different occupations in Figure \ref{fig:occupation}. The figure shows that the $0.75$ quantiles of association scores for computer programmers and pharmacists are higher than the others by a large margin, indicating that these occupations are more strongly associated with men. Conversely, the mean association scores for elementary school teachers, librarians, announcers, and chemists are negative, indicating that these occupations are more strongly associated with women. The association score for chef and police is neutral, suggesting that there is insufficient evidence to establish a stereotype.

\begin{figure}[t]
    \centering
    \includegraphics[width=\linewidth]{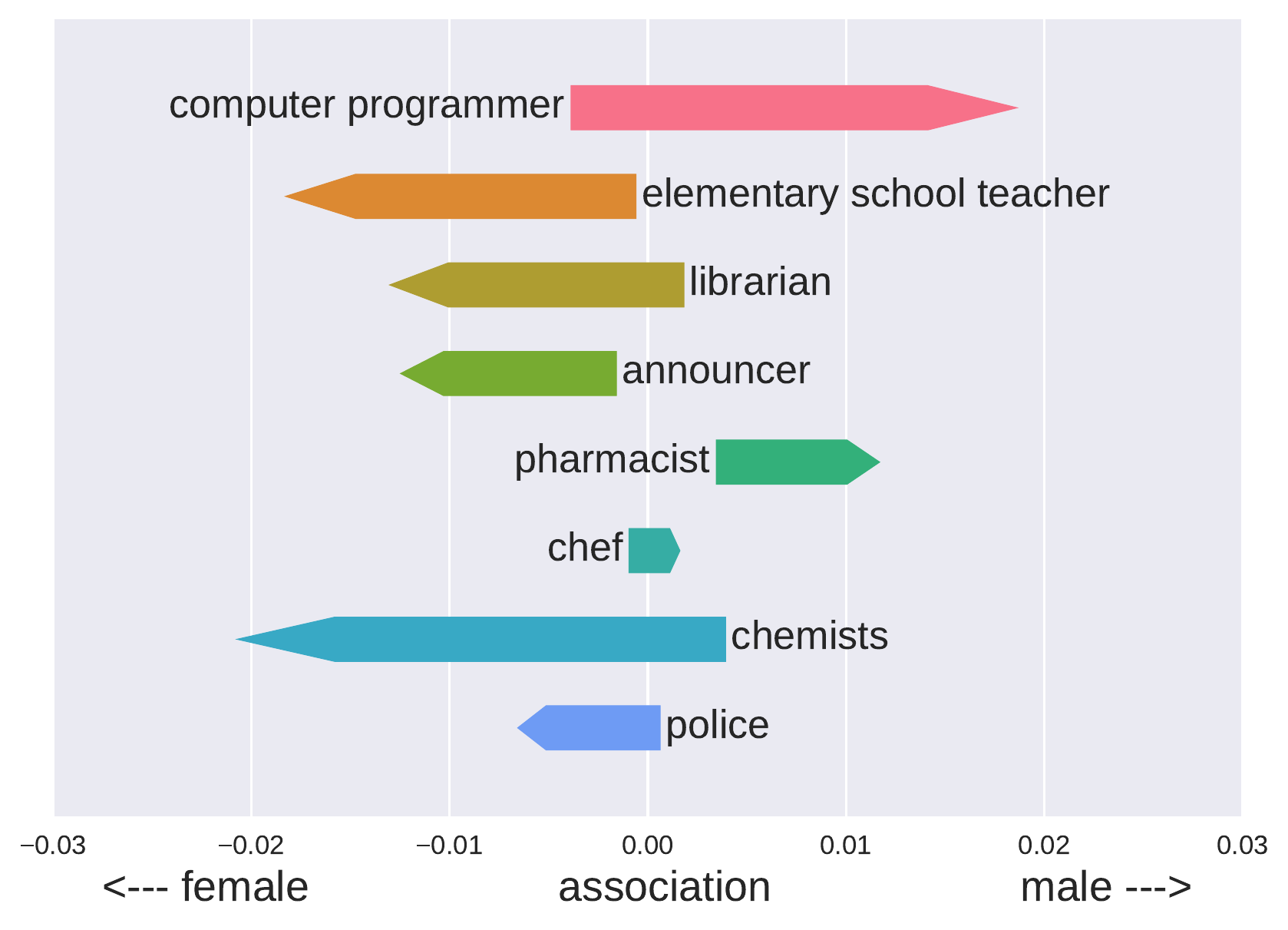}
    \caption{\textbf{Stereotype amplification.} For each occupation, we compare the association scores for generated images to the association scores for the text prompts. The association scores for the text prompts are represented by the tails of the arrows, and the association scores for the images are represented by the heads of the arrows.}
    \label{fig:bias-amplification}
\end{figure}

\begin{table}[!ht]
    \centering
    \resizebox{\linewidth}{!}{
    \begin{tabular}{ccc}
    \toprule
        Concept & Attribute & Score \\
    \midrule
        Flowers & Pleasant \vs Unpleasant & 1.00\\
        Insects & Pleasant \vs Unpleasant & 0.15\\
        Musical Instrument & Pleasant \vs Unpleasant & 0.90\\
        Weapon & Pleasant \vs Unpleasant & 0.05 \\
    \midrule
        Science & Male \vs Female & 0.75\\
        Arts & Male \vs Female & 0.30 \\
        Careers & Male \vs Female & 0.75\\
        Family & Male \vs Female & 0.40\\
    \bottomrule
    \end{tabular}
    }
    \caption{\textbf{Human evaluation results.} For each pair of concept and attributes, we report the fraction of images that are chosen as being more closely associated with pleasant or male attributes. We find out that the machine-rated association scores can properly represent human's perceptions.}
    \label{tab:human_eva}
\end{table}
\subsection{Stereotype Amplification}
Do images generated by the diffusion model amplify the implicit stereotypes in the textual representations used to guide image generation? Specifically, we examine occupational images and calculate the association scores between the text prompts by substituting the text embeddings of CLIP into Eq.~(\ref{eq:asc}) and Eq.~(\ref{eq:differential-association}). We then compare these associations for text prompts to the associations for the generated images to investigate whether the biases are amplified. 

Figure~\ref{fig:bias-amplification} demonstrates the stereotype amplification between text prompts and generated images. For each occupation, we use an arrow to represent the change of associations on the axis of gender. We observe that the associations are amplified on a large scale for most occupations. In particular, the textual association between a computer programmer and gender is only $-0.0039$ but enlarged to $0.0186$ for images. Similar amplifications are observed for elementary school teachers, librarians, and chemists. For the occupation of chef, the association of text prompts is skewed towards females, while the association of images is skewed towards males.

\subsection{Comparison to Human Evaluation}
We recruit university students to evaluate the generated images and compare how the perceptions of human differ with the machine-evaluated association scores.  Specifically, for each set of concepts, we ask three student participants to view 20 images generated with neutral prompts and choose which valence or stereotypical attribute is more closely associated. We report the fraction of images that are chosen as being more closely associated with pleasant or male attributes.
As shown in Table \ref{tab:human_eva}, the human's preference of association aligns with the strength of our association scores. For flowers \vs insects and musical instruments \vs weapon, humans mostly prefer to associate flowers and musical instruments with pleasant while insects and weapons with unpleasant. For science \vs arts and career \vs family, we find that the significance of the bias is reduced. The Kendall's $\tau$ coefficient between the machine-evaluated and human-rated scores is 0.55, indicating that the association scores can properly represent human's perceptions.

\section{Discussion}
Our bias test was applied to testing biases of images generated by the state-of-the-art text-to-image generative model, associated with valence and gender attributes of variation of concepts such as careers, religions, skin tone, etc. In the example of the valence test for images generated for Straight \& Gay concepts, we observed a significant bias of pleasant attitudes towards people with straight sexual orientation and unpleasantness towards people with gay sexual orientation; the findings successfully mirrored the acknowledged human biases. Similar to the Stable Diffusion example we selected in our work, the proposed bias test can be applied to other generative models with the experiment in resemblance to quantify existing implicit biases.

The proposed Text-to-Image Association Test is a principal approach for measuring the complex implicit biases in image generations. The primary result illustrates the valence and stereotypical biases across various dimensions, ranging from morally neutral to demographically sensitive, in a state-of-the-art generative model at different scales. The presented research adds to the growing literature on AI ethics by highlighting the complex biases present in AI-generated images and serves as a caution for practitioners to be aware of these biases.

\section{Limitations}
Our work has some limitations. Although we use the same verbal stimuli in the previous IAT tests for creating text prompts, it is very likely that some stimuli that can represent the concepts are underrepresented. The approach we adopted for comparing the images' distance might be biased as well. The current bias test procedure applies the visual encoder of OpenAI's CLIP model to measure the distance between images. However, it is unclear whether the image encoder may inject additional biases into the latent visual representations. 

\section*{Ethics Statement}
The scope of this work is to provide a principal procedure for measuring the implicit valence and stereotypical biases in image generations. The experiments conducted involve generating images that pertain to demographic groups, and all images were generated in compliance with the terms of service and guidelines provided by the stable diffusion's license. The AI-generated images are used solely for research purposes and no identities are explicitly attributed to individuals depicted in the images. People's names are used to generate images. We justify that these are common American names publicly accessible, and do not contain any information that can uniquely identify an individual.

\section*{Acknowledgement}
We thank the anonymous reviewers for their constructive comments. This work is primarily supported by X. Wang's startup fund. J. Wang and Y. Liu are also partially supported by the National Science Foundation (NSF) under grants IIS-2143895 and IIS-2040800.

\bibliography{anthology,custom}
\bibliographystyle{acl_natbib}

\appendix

\section{Additional Experiment Details}\label{app:additional-experiment-detail}
We show the detailed verbal stimuli for all the 8 bias tests in Table~\ref{tab:concept-detailed}.

\begin{table*}
    \centering
\resizebox{0.8\linewidth}{!}{
\begin{tabular}{l p{0.7\linewidth}}
\toprule
Concept & Verbal Stimuli \\
\midrule
Flowers &  aster, clover, hyacinth, marigold, poppy, azalea, crocus, iris, orchid, rose, bluebell, daffodil, lilac, pansy, tulip, buttercup, daisy, lily, peony, violet, carnation, gladiola, magnolia, petunia, zinnia. \\
Insects & ant, caterpillar, flea, locust, spider, bedbug, centipede, fly, maggot, tarantula, bee, cockroach, gnat, mosquito, termite, beetle, cricket, hornet, moth, wasp, blackfly, dragonfly, horsefly, roach, weevil. \\
Musical Instruments & bagpipe, cello, guitar, lute, trombone, banjo, clarinet, harmonica, mandolin, trumpet, bassoon,
drum, harp, oboe, tuba, bell, fiddle, harpsichord, piano, viola, bongo, flute, horn, saxophone, violin. \\
Weapon & arrow, club, gun, missile, spear, axe, dagger, harpoon, pistol, sword, blade, dynamite, hatchet, rifle, tank, bomb, firearm, knife, shotgun, teargas, cannon, grenade, mace, slingshot, whip.\\
European American & Adam, Chip, Harry, Josh, Roger, Alan, Frank, Ian, Justin, Ryan, Andrew, Fred, Jack,
Matthew, Stephen, Brad, Greg, Jed, Paul, Todd, Brandon, Hank, Jonathan, Peter, Wilbur, Amanda, Courtney, Heather,
Melanie, Sara, Amber, Crystal, Katie, Meredith, Shannon, Betsy, Donna, Kristin, Nancy, Stephanie, Bobbie-Sue, Ellen,
Lauren, Peggy, Sue-Ellen, Colleen, Emily, Megan, Rachel, Wendy (deleted names in italics). \\
African American & Alonzo, Jamel, Lerone, Percell, Theo, Alphonse, Jerome, Leroy, Rasaan, Torrance, Darnell,
Lamar, Lionel, Rashaun, Tvree, Deion, Lamont, Malik, Terrence, Tyrone, Everol, Lavon, Marcellus, Terryl, Wardell,
Aiesha, Lashelle, Nichelle, Shereen, Temeka, Ebony, Latisha, Shaniqua, Tameisha, Teretha, Jasmine, Latonya, Shanise,
Tanisha, Tia, Lakisha, Latoya, Sharise, Tashika, Yolanda, Lashandra, Malika, Shavonn, Tawanda, Yvette (deleted names
in italics). \\
light skin & light-skinned person, light-skinned girl, light-skinned woman, light-skinned women, light-skinned boy, light-skinned man, light-skinned men, light-skinned family, light-skinned community. \\
Dark skin & dark-skinned person, dark-skinned girl, dark-skinned woman, dark-skinned women, dark-skinned boy, dark-skinned man, dark-skinned men, dark-skinned family, dark-skinned community. \\
Straight & straight person, straight girl, straight woman, straight women, straight boy, straight man, straight men, straight family, straight community. \\
Gay & gay person, gay girl, gay woman, gay women, gay boy, gay man, gay men, gay family, gay community. \\
Judaism & synagogue, torah, jew, judaism. \\ 
Christianity & church, bible, christian, christianity. \\
Career & executive, management, professional, corporation, salary, office, business, career. \\
Family & home, parents, children, family, cousins, marriage, wedding, relatives. \\
Science & science, technology, astronomy, math, chemistry, physics, biology, geology, engineering. \\
Arts & poetry, art, history, humanities, English, philosophy, music, literature. \\
\midrule
Pleasant & caress, freedom, health, love, peace, cheer, friend, heaven, loyal, pleasure, diamond, gentle, honest, lucky, rainbow, diploma, gift, honor, miracle, sunrise, family, happy, laughter, paradise, vacation.\\
Unpleasant & abuse, crash, filth, murder, sickness, accident, death, grief, poison, stink, assault, disaster, hatred, pollute, tragedy, bomb, divorce, jail, poverty, ugly, cancer, evil, kill, rotten, vomit. \\
Male &  male, man, boy, brother, son.\\
Female & female, woman, girl, sister, daughter. \\

\bottomrule
\end{tabular}}
\caption{\textbf{Verbal stimuli for each of the concepts and attributes.}}
\label{tab:concept-detailed}
\end{table*}

\end{document}